\pdfoutput=1

\documentclass[11pt]{article}

\usepackage[review]{acl}

\usepackage{times}
\usepackage{latexsym}

\usepackage[T1]{fontenc}

\usepackage[utf8]{inputenc}

\usepackage{microtype}

\usepackage{color}
\usepackage[colorinlistoftodos]{todonotes}
\usepackage{xcolor}

\title{Parsimonious Argument Annotations for Hate Speech Counter-narratives}

\author{
  Damián A. Furman$^{1}$, Pablo Torres$^{2}$, Jose A. Rodriguez$^{2}$, Lautaro Martínez$^{2}$,\\
  Laura Alonso Alemany$^{2}$, Diego Letzen$^{2}$ and Maria Vanina Martinez$^{1}$ \\
  \\
  \small
  1 - Universidad de Buenos Aires; 2 - Universidad Nacional de Córdoba\\}

\begin{document}
\maketitle
\begin{abstract}
We present an enrichment 
of the Hateval corpus of hate speech tweets~\cite{basile-etal-2019-semeval} aimed to facilitate automated counter-narrative generation. Comparably to previous work~\cite{chung2019conan}, manually written counter-narratives are associated to tweets. However, this information alone seems insufficient to obtain satisfactory language models for counter-narrative generation. That is why we have also annotated tweets with argumentative information based on~\citet{Wagemans2016ConstructingAP}, that we believe can help in building convincing and effective counter-narratives for hate speech against particular groups. 

We discuss adequacies and difficulties of this annotation process and present several baselines for automatic detection of the annotated elements. Preliminary results show that automatic annotators perform close to human annotators to detect some aspects of argumentation, while others only reach low or moderate level of inter-annotator agreement.

\end{abstract}

\section{Introduction and Motivation}

With the rapid growth of social media platforms, hate speech and hate messages that were already present in our society found a way to spread faster and increase their reach.\footnote{The International Convention on the Elimination of all Forms of Racial Discrimination (CERD), understands hate speech as "a form of speech directed at others that rejects basic human rights principles of human dignity and equality and seeks to degrade the position of individuals and groups in society's esteem". (United Nations Strategy and Plan of Action on Hate Speech: Detailed Guidance on Implementation for United Nations Field Presences, 2020)

} 
The predominant strategy adopted so far to counter hate speech in social media is to recognize, block and delete these messages and/or the users that generated it. This strategy has two main disadvantages. The first one is that blocking and deleting may prevent a hate message from spreading, but does not counter its consequences on those who were already reached by it. The second one is that there is no place for subtleties or shades while defining hate speech: it must be done as a binary classification because the consequence of that classification is binary. This can generate accusations of overblocking or censorship and not just because of errors in automated systems, which could be arguably improved. On the contrary, blocking seems to be much too simplistic an approach to deal with the inherent complexity of hate speech.




An alternative to blocking that has been gaining attention in the last years, is to "\textit{oppose hate content with counter-narratives (i.e. informed textual responses)}"~\cite{Benesch2014,chung2019conan}\footnote{No Hate Speech Movement Campaign: http://www.
nohatespeechmovement.org/}. 
However, automating the generation of counter-narratives is no easy task. Recent approaches based on neural sequence-to-sequence architectures~\cite{tekiroglu2020guerini} provide an interesting methodology to obtain generators of narratives inferred from existing pairs of hate speech -- counter-narrative. But for these approaches to perform adequately, huge amounts of examples are needed, and they are not currently available for this task. 
That is why we propose to boost the performance of approaches based on text alone by providing an argumentative analysis of these texts that facilitates identifying relevant elements that can be of use to generate counter-narratives.

We expect argumentative annotations to provide insight into the usefulness of different aspects of argumentation for this task. At this point of our work, our main research question is to \textbf{determine the difficulty to identify different aspects of argumentation in hate speech}. We want to assess difficulty both for human annotators and for automatic models. 

In this paper we present the Argumentation Structure For Counter-Narrative Generation (ASFoCoNG), a protocol to annotate domain-specific argumentative information and an annotated dataset to train automatic classifiers. These annotations are an adaptation of~\citet{Wagemans2016ConstructingAP}'s proposal, which we believe provides an adequate starting point for our purpose.

We also present some preliminary experiments that show promising results for the automatic detection of some aspects of argumentation, like Justifications or Collectives, while highlighting the highly interpretative, subjective nature of others, like Pivots or Properties.

The rest of the paper is organized as follows. In the following section, we discuss relevant work on corpora annotated for counter-narratives and argument annotation schemes. Then we describe our annotation scheme, 
and we finish by providing preliminary results for automatic identification of argumentation and insights gained from difficulties and discrepancies between annotators.

\section{Relevant Work}

\subsection{Annotated counter-narrative corpora}

As far as we are aware of, there are two datasets containing hate-speech with their corresponding response pairs that are publicly available. 

The CONAN dataset~\cite{chung2019conan} contains 4078 Hate Speech-Counter Narrative original pairs manually written by NGO operators, in three languages: English, French and Italian. Data was augmented using paraphrasing and translations between languages to 15024 pairs of hate speech -- counternarrative.
Unfortunately, this dataset is
not representative of the language in social media. 

Qian et al.~\cite{DBLP:journals/corr/abs-1909-04251} have released a dataset with conversations from 5020 reddit posts with 22324 comments and 11825 conversations from Gab with 33776 comments. Mechanical Turk workers labelled specific comments on each conversation as hateful or not and then wrote responses to them. 21747 responses were written for the Gab dataset and 7641 were written for Reddit. Though this dataset adds a great value by contextualizing hate speech inside a conversation, the response to those messages are focused mainly on moderation of the debate (e.g. warnings against using offensive language) instead of generating proper counter-narratives against the hate message itself.

None of the aforementioned datasets includes any additional annotated information apart from the hate message (in the case of~\cite{DBLP:journals/corr/abs-1909-04251}, identified inside a conversation) and its response.
These datasets are well-suited for neural sequence to sequence approaches~\cite{devlin-etal-2019-bert},  which take one text string and output another, in this case, they take a hate narrative and output a counter-narrative. However, these approaches require huge amounts of text to achieve good performance.



\subsection{Argument annotation}

There are many datasets with argument annotations considering different aspects, but to the best of our knowledge there are only two annotating arguments on Twitter: DART~\cite{bosc-etal-2016-dart} and a subset of DART's extension made by \cite{dusmanu-etal-2017-argument}. DART relies on Argumentation Theory~\cite{rahwan2009argumentation} to find relationships between tweets as a single unit, considered to be arguments within an Abstract Argumentation Framework~\cite{Dung95}. Tweets are considered as argumentative if they express opinion or claims showing stance about a particular topic, and then they are defined according to how they interact with other tweet-arguments. \cite{dusmanu-etal-2017-argument} extended the \textit{\#Grexit} subset of DART (987 tweets) with another 900 labeled for argument detection and added labels for two new tasks, factual arguments recognition and source identification.


An abstract argumentation framework is incapable of considering the inner structure of an argument, and therefore is not useful to detect argumentative information that can later be used to counter that argument.
We aim to construct a system that is able to identify argumentation within a single tweet and extract relevant information that can be used to generate counter-narratives.

\subsubsection{Argument Schemes}

Argument or argumentation schemes are "\textit{abstractions substantiating the inferential connection between premise(s) and conclusion in argumentative communication}"~\cite{annotatingArgumentSchemesVisser}. They are "patterns of informal reasoning"~\cite{walton_reed_macagno_2008} that "represent forms of argument that are widely used in everyday conversational argumentation"~\cite{macagnoArgSchemes2018}. 

Intuitively, argument schemes provide adequate information to create a counter-narrative. For example, if an \textit{authority} scheme is identified, a response of the kind "\textit{how reliable is this authority?}" or "\textit{is this authority an expert on the field in discussion?}" could be generated to serve the purpose of undermining the message.

\subsubsection{Walton's Schemes}


\newcite{walton_reed_macagno_2008} schemes follows a 'bottom-up' approach that begins with real examples of arguments that can be clustered together into a scheme, and then those schemes can again be clustered together into more general classifications of schemes. 
Many authors have adapted some of Walton's schemes to their specific domain or purpose or proposed their own with their corresponding critical questions~\cite{ATKINSON20181}~\cite{kokciyan2018}.

The main drawback of this proposal is that the inventory of scheme is very profligate, with many different schemes. No formal guidelines exist for argument type identification, 
making the task very difficult both manually and automatically. 

\subsubsection{The Periodic Table of Arguments}

\newcite{Wagemans2016ConstructingAP} proposes an analytic approach to argument schemes, aimed to obtain the core schemes proposed by \newcite{walton_reed_macagno_2008} with a smaller number of categories. 
This is a characteristic that we find particularly useful for building a simple system that is easy to annotate without an enormous effort and achieving a high level of agreement between human annotators, which leads us to expect higher reproducibility in inferred models. Moreover, an analytic approach allows determining which aspects of argumentation are more feasible to detect automatically, and which may be more useful to guide counter-narrative generation.

The Periodic Table of Arguments (PTA) is a factorial typology of arguments that 
defines them as a unique combination of three basic characteristics~\cite{fourArgumentForms2019}: 
\begin{enumerate} \itemsep -0.1cm
    \item \textbf{first order or second order}. All arguments under Wagemans's system will have a premise and a conclusion, with a common term that transfers acceptability from one to the other. If this common term is explicit and no reconstruction of the argument is necessary to find it then it is a first order argument. If a reconstruction is needed (usually by transforming a premise into the subject of another premise with the predicate "is true") then it is a second order argument.
    \item \textbf{predicate or subject}. If the common term is in the subject of the propositions making the premise and the conclusion, 
    then it is a subject argument, otherwise, it is a predicate argument. 
    \item \textbf{policy, fact or value}. The conclusion and premise of an argument can be labelled each as a statement of policy (the speaker mandates that something should be done), of value (the speaker issues an opinion), or of fact (the speaker conveys a proposition as a true fact).
\end{enumerate}


We adapted Wagemans's proposal to hate speech counter-narrative generation. Our main focus is not discriminating types of arguments, but identifying elements that can help to build better counter-narratives.

\section{Annotation of argumentative information in hate speech in twitter}

A corpus obtained from Twitter, more specifically a hate speech corpus, is full of incomplete, incoherent and syntactically ill formed sentences. Many argumentative hate tweets are based on assumptions justified by prejudice or context information that is difficult to recover. This means that in many cases, it is difficult to rebut them from the perspective of formal deductive logic. 
We believe that an informal logic based approach could be useful to capture argument as they occur in this context.


The goal of our proposed annotation is to provide meaningful information with  domain-specific categories that can be used to generate counter-narratives, both at annotation time by suggesting the annotator how to write the counter-narratives based on the previously labelled components and at prediction time using machine learning. 


We created an annotation manual\footnote{\url{https://docs.google.com/document/d/1AZ1hbrxhZ-n6Lzb9fuaBZd8kAmuLcOlahG6I9dUGz_g}} describing the process to annotate each hate speech tweet, with 
the following steps: \textbf{(1)} determine whether the tweet is argumentative; if it is argumentative, \textbf{(2)} identify premise and conclusion, \textbf{(3)} identify the collective to whom the hate speech is directed, and the properties assigned to them, 
\textbf{(4)} identify an element that transfers reasoning from premise to conclusion (we call this the \textit{pivot}), \textbf{(5)} determine whether the premise and conclusion are \textit{policy}, \textit{fact} or \textit{value} statements, and \textbf{(6)} write counter-narratives.

A trained philosopher annotated 80\% of the tweets. A trained computer scientist annotated 20\% of the tweets, and they both annotated 80 tweets to assess reproducibility of annotation across judges. 





\begin{figure*}
    \centering
    \includegraphics[width=.95\linewidth,trim={0 2.5cm 0 0.5cm} , clip]{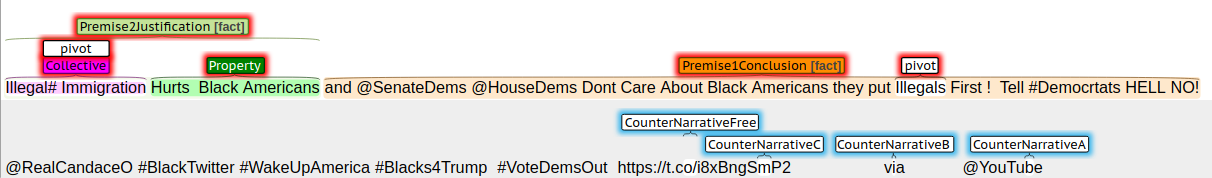}
    \caption{\label{fig:argumentative} Example of argumentative hate tweet with marked Justification and Conclusion (of type \textit{fact}), Collective, Property and Pivot.}
\end{figure*}

\subsection{Argumentative or non-argumentative}

We consider a tweet to be argumentative if there is "a statement that is put forward in order to support another statement whenever the acceptability of the latter is in doubt \cite{Wagemans_2019}".
Following Wagemans's, "an argument (...) consist(s) of two statements, namely a conclusion – the statement that is doubted – and a premise". So if it is possible to divide the tweet in these two segments, the tweet is considered as argumentative. 

Examples of non-argumentative tweets are: exhortations to some action without justification, insults, name callings, support for a particular policy or description of facts without an explicit conclusion. 
An example of a non-argumentative tweet follows:
\vspace{-0.03cm}
\begin{quote}
    \texttt{No to \#EU migrant camps in Libya, PM al-Serraj https://t.co/hlsSOU73lQ https://t.co/SxaU7QhDTn}
\end{quote}


\subsection{Reconstruct the most salient component}

If the tweet is argumentative, this means that it is possible to identify a conclusion and a justification. All argumentative tweets are labeled with both and there is only one of each, although they can be separated in many non-contiguous parts inside the tweet. The annotators were instructed to try to fit the most of the example inside this two components, leaving out only hashtags indicating topics, links or user mentions or non-relevant words or information. Justifications may be arguments themselves, having their own inner structure involving different premises, but everything is going to be considered as justification without further distinction. For the purpose of building counter narratives, we are only interested in capturing the main standpoint that the user wants to gain acceptability for.

This classification is based on a simplified version of Wagemans's Periodic Table of Arguments,  without considering the subject-predicate structure of each component. Visser et al.~\cite{annotatingArgumentSchemesVisser} warned about how PTA presupposes that premises and conclusions of arguments consist of complete categorical propositions comprising a clear subject-predicate structure and the problems that arise in the US presidential debate corpus as a result of "interruptions, corrections, and general obscurity" that causes the transcription of the speech to be "incomplete or not syntactically well formed". We detected the same problem to be even more severe in social media noisy user generated examples. Moreover, here we are not interested on classifying the arguments but rather on identifying the basic argumentative characteristics of the message.

\subsection{Collective and Properties}

All hate messages are directed towards a specific group 
by definition.
Usually, the content of the message is to associate this group with a negative property or an undesirable action or consequence. If this property\footnote{We consider as property anything that is associated with the targeted community, whether it is an adjective, a consequence, an action, etc.} is explicit, we label it.


\subsection{Pivot}


In order to be able to transfer reason, a justification and a conclusion must have a common element: we call this the pivot. 
We identify pivots as two sequences of words, one for each premise, that can be related to the element that those premises have in common. 
It is important to note that this relation is generally not unique and very deep in the layers of meaning of the argument, therefore it may be represented by quite different sequences of words. Whenever this element is explicit, we annotate it.

The pivot holds a relation with Wagemans's classification of first and second order classification of arguments. If an argument is considered first-order, it means that the common element between premises must be explicit (because, by definition, it must be either of the form A ix X because A is Y or B is Z because C is Z). If it is a second-order argument, there might still be an explicit pivot though this is not warranted.

\subsection{Types of proposition}

Wagemans proposes "a characterization of the types of arguments based on the combination of the types of propositions they instantiate"~\cite{Wagemans2016ConstructingAP}. These types are taken from debate theory 
where three distinctions are made: (1) the proposition of policy (P), (2) the proposition of value (V) and (3) the proposition of fact (F). We label our propositions using the same types and add to our annotation manual different guidelines on how to recognize each one: a policy proposition is a mandate often expressed as orders or imperatives, or actions that need to be accomplished in the public domain. Distinction between fact and value propositions was reported to be significantly more difficult to make. We establish as general criteria that a proposition to be labeled as value must have explicit markers of the speaker being involved in the assertion expressed, like opinionated adjectives, verbs of thought, and others. Otherwise, the premise is considered as fact. 

\subsection{Counter-Narratives}
Each tweet is associated to specific counter-narratives, divided in four different types. Each type defines a different guideline used to build the counter-narrative based on the different components previously annotated. 
Annotators were suggested to try to write at least one counter-narrative of each type but only if they came naturally from the given hints, otherwise they could leave it blank. 


\subsubsection{Negate Relation Between Justification And Conclusion (type A)}

This type of counter-narrative is based on attacking the core of the user's reasoning by negating the implied relation between the justification and the conclusion. All strategies based on attacking this implied relation fall under this category.

\subsubsection{Negate association between the targeted collective and the property given to them  (type B)}

This type of counter-narrative aims to attack the relation between the property, action or consequence that is fallaciously trying to be assigned to the targeted group. It requires that the property were explicitly mentioned on the original tweet, otherwise, it is left blank.

\subsubsection{Attack the justification premise based on it is type  (type C)}

Reacting to hate speech cannot be solely based on detecting faults in reasoning, but it needs to question the values supported by hate speech. This type of counter-narrative aims to attack the justification based on its type. If the justification is a fact, then the fact must be put into question or sources must be asked to prove that fact. If it is of type “value”, the characteristic of the premise of being an opinion must be highlighted and the speaker’s opinion itself must be relativized as a xenophobous opinion. If it is a “policy” a counter or opposite policy must be answered.

\subsubsection{Free Counter-Narrative type D}

If the annotator recognize a counter-narrative that does not fit in any of the previous three types she is encouraged to write it down under this fourth type. This fourth type includes all counter-narratives that are not part of any of the other three.

\begin{table*}[ht!]\centering
    \vspace{-0.5cm}
    \begin{tabular}{l|ccc|cc|c|cc}
        \hline
                          &   Collect.   & Prop.          &    Pivot       &   Justif.         &  Conc. & Arg. & Type Conc. & Type Just.\\
        \hline

        Cohen's $\kappa$    &  .67 & .53 & .56 & .71 & .69 & .85 & .65 & -.05\\
        \hline
        \hline
        human annotator F1   &  .68 & .58 & .58 & .84 & .77 & .95 & .74 & .44\\
        automatic annotator F1   &  .78 & .52 & .32 & .79 & .52 & .53 & .73 & .64\\
        \hline
    \end{tabular}
    \caption{Agreement scores for 80 tweets labeled by two annotators (a philosopher and a computer scientist) average F1 between annotators and F1 obtained by the best average automatic classifier, RoBERTa.}
    \label{tab:agreement}
    \vspace{-0.5cm}
\end{table*}

\section{The ASFoCoNG corpus\footnote{A brat server with the annotated dataset can be found at https://github.com/DamiFur/brat-annotations-argumentation-schemes. The repository also contains usefull scripts for calculating agreement and running several experiments}}

We applied the above guidelines to a subset of the HatEval 2019 corpus~\cite{basile-etal-2019-semeval} containing tweets in English and Spanish. We discarded tweets labeled as "aggressive", consisting mostly of abusive language (name callings, insults, exhortations to action and other types of attacks) and tweets targeted against specific individuals, because they were almost exclusively non-argumentative. We also filtered all tweets against women, since we noticed that the majority of them were also based on abusive language and were also non-argumentative. This left us with 970 tweets in English and 196 tweets in Spanish with a total of 25413 words in English and 5436 in Spanish.

From the 970 tweets in English, 245 (25.3\%) were labeled as Non-Argumentative while the other 725 (74.7\%) were considered argumentative. From these argumentative tweets, 422 (58.2\%) have an explicit Collective (593 words) and Property (2092 words) and 327 (45.1\%) have a pivot (875 words). All argumentative tweets are labeled with a Conclusion (7306 words) and a Justification (11708 words) by definition. 267 (36.8\%) conclusions were labeled as "facts", 41 (5.7\%) as "values" and 417 (57.5\%) as "policies", while 675 (93.1\%) Justifications were labeled as "fact", 23 (3.2\%) as "value" and 27 (3.7\%) as "policy". Of the 725 argumentative tweets, 697 of them have a counter-narrative of type A; 339 have a counter-narrative of type B; 653 have a counter-narrative of type C; and 9 have a counter-narrative of type D.

The Spanish corpus has 52 (26.5\%) Non-Argumentative and 144 (73.5\%) Argumentative tweets. From these argumentative tweets, 88 (61.1\%) are labeled with a Collective (132 words) and Property (365 words) and 54 (37.5\%) have a pivot (135 words). Conclusions consist of 1844 words, while Justifications have 2683 words. Also, 81 (56.2\%) Conclusions are labeled as "fact", 23 (16\%) as values and 40 (27.8\%) as policies and 139 (96.5\%) Justifications are labeled as "fact", 2 (1.4\%) as "values" and 3 (2.1\%) as "policies". Of the 144 argumentative tweets, 140 have a counter-narrative of type A; 88 have a counter-narrative of type B; 128 have a counter-narrative of type C; and only 1 has a counter-narrative of type D.

Annotation was done using the brat annotation tool \cite{stenetorp-etal-2012-brat}.

\subsection{Inter-annotator agreement}


We carried out an inter-annotator agreement evaluation to assess the reproducibility of the annotations, and to address the question of which of the proposed categories can be successfully systematized.

To calculate inter-annotator agreement, both annotators labeled 80 tweets (8\% of the corpus) of the English portion of HatEval. Then, per-category agreement was calculated with Cohen's $\kappa$~\cite{cohen1960} and also F1, considering one of the annotators as the ground truth and the other as the one to evaluate. Agreement was calculated in a per-tweet basis for the Argumentative vs. Non-Argumentative category using a binary label, and for the Type of Conclusion and Justification categories, using one label with three possible values representing \textit{fact}, \textit{value} and \textit{policy}. For all other categories, agreement was calculated in a per-word basis with a binary label assigned to each word, marking whether it belongs to the category or not.

Table~\ref{tab:agreement} shows inter-annotator agreements and the performance of the best automatic annotator. We can see that human annotators can reach an substantial level of reproducibility, around $\kappa=.7$ for Collective, Justification, Conclusion, Type of Conclusion and the distinction between Argumentative or non-Argumentative tweets. 

In contrast, the Property and Pivot categories present moderate levels of inter-annotator agreement, and the Type of Justification presents no agreement at all. Some examples of disagreement between annotators can be seen in Annex~\ref{app:examples}.

The disagreement over the Type of Justification seems to be due to the fact that some \textit{values} are very hardly distinguishable from \textit{facts}. Figure~\ref{fig:disagreement_type_justification} shows and discusses an example illustrating this difficulty.

When inspecting examples of disagreement between annotators for Pivot, as the one shown in Figure~\ref{fig:disagreement_pivot}, we found that in many cases both annotations could be considered as accurate. The relation that the pivot intends to capture between Justification and Conclusion is not necessarily logical or formal, but informal, and there may be more than one possibility for annotators to tag. Furthermore, since the relation is very deep in the layers of meaning, annotators may interpret it as signalled by different surface features, and as a consequence they may tag different sequences of words while considering the same relation.

Lastly, it can be seen that not all categories that can be systematically identified by humans can be successfully identified by automatic annotators. Indeed, the best automatic annotator presents low performance for the distinction between Argumentative and Non-Argumentative, which is the category with highest inter-annotator agreement. In contrast, for Conclusion and Type of Justification, performance is better than human agreement, showing that the model could identify some systematicities that humans were not applying in their decisions. We will be studying the behavior of the models to try to capture such systematicities and integrate them to improve annotation guidelines or to gather more examples to enhance our corpus. For Property and Pivot, with low inter-annotator agreement, the performance of the automatic annotators was also worse than  for other elements. Particularly for Pivot, the performance of the automatic model was also significantly worse than a human annotator.

\section{Automatic classifiers}

We trained different kinds of automatic classifiers in order to assess the difficulty for automatic identification of different aspects of argumentation in the annotated corpus. We only conducted experiments for the English portion of the dataset, because the Spanish portion is too small for training.


We used 79\% of the English portion of the dataset as training (770 tweets), 10.5\% as development (100 tweets) and 10.5\% as test (100 tweets). For each set of hyperparameters used, we trained three models using different random tweets for each partition, always respecting this proportion. We report the average of these three models F1, Precision and Recall and the standard deviation for F1 scores. For the task of predicting the type of premise, we report the F1 macro average between the three possible labels. For all other binary classification tasks, we report the F1 for the target class, that is, F1 of the words labelled as Property, Justification, etc.

To be able to automatically re-create the annotation process, models should be provided with the same information that an annotator would have at the correspondent stage of annotation: for Collective, she would know if there is an explicit Property and for Pivot and the types of the two premises, she would already know the Justification and Conclusion. For these four categories, we ran all experiments with and without adding this information to assess how important is the order in the annotation process and the knowledge of previous annotations.

We describe now the different models we trained and evaluated.







\subsection{Logistic Regression}

As a baseline, we used Scikit-learn implementation of Logistic Regression~\cite{scikit-learn} and modeled the input data with two different approaches: using a bag-of-words and using contextual embeddings generated by extracting the last layer of the RoBERTa model we also evaluated. We tried four different values for inverse of regularization strength: 1.0, 0.1, 0.2 and 0.5. 

\subsection{RoBERTa}
RoBERTa~\cite{liu2019Roberta} is a pre-trained language model based on a Transformers architecture similar to BERT~\cite{devlin2018bert} but trained with more data. 
For this model we tried 8 different values for learning rate (1e-05, 2e-05, 5e-05, 8e-05, 1e-04, 2e-04, 5e-04 and 8e-04) and used the development dataset to implement early stopping with a maximum of 10 epochs.

\subsection{BiLSTM-CNN-CRF}

We used the BiLSTM-CNN-CRF implementations by UKP Lab~\cite{reimers} for sequence tagging, composed of a BiLSTM with a Conditional Random Field linear classification layer at the end and char embeddings obtained by a 1D-CNN used to enrich the input. Besides the standard configuration, we also tried switching the CRF layer for a Softmax and using three different word embbedings as input for the model: FastText \cite{bojanowski2016enriching}, Komninos \cite{komninos-2016} and contextual embeddings generated using ELMO~\cite{elmo}.




\subsection{Results}

\begin{table*}[ht!]
   \begin{tabular}{|l|ccc|ccc|ccc|ccc|} 
        \hline
                        & \multicolumn{3}{c|}{LR}    & \multicolumn{3}{c|}{LR w/embed}  & \multicolumn{3}{c|}{BiLSTM+CRF} & \multicolumn{3}{c|}{RoBERTa}   
        \\\hline
                        & F1 & Pr & Rec & F1 & Pr & Rec & F1 & Pr & Rec & F1 & Pr & Rec \\ 
        \hline
        Arg./Non-Arg. & .0 & .0 & .0 & .51\small{$\pm$.13} & .50 & .52 & \textbf{.54}\small{$\pm$.10} & .55 & .53 & .53\small{$\pm$.08} & .50 & .57\\ 
        \hline
        Justification & .50\small{$\pm$.004} & .46 & .56 & .68\small{$\pm$.00} & .62 & .75 & .72\small{$\pm$.08} & .68 & .78 & \textbf{.79}\small{$\pm$.04} & .76 & .82\\ 
        Conclusion & .35\small{$\pm$.03} & .27 & .49 & .48\small{$\pm$.01} & .37 & .70 & \textbf{.58}\small{$\pm$.08} & .57 & .59 & .52\small{$\pm$.03} & .58 & .44\\ 
        Type of Just. & .32\small{$\pm$.004} & .31 & .33 & .32\small{$\pm$.00} & .31 & .33 & - & - & - & \textbf{.36}\small{$\pm$.06} & .34 & .32\\ 
        Type of Conc. & .25\small{$\pm$.01} & .20 & .33 & .43\small{$\pm$.03} & .43 & .44 & - & - & - & \textbf{.50}\small{$\pm$.05} & .48 & .51 \\ 
        Collective    & .10\small{$\pm$.02} & .06 & .41 & .52\small{$\pm$.05} & .37 & .85 & .52\small{$\pm$.07} & .64 & .43 & \textbf{.59}\small{$\pm$.02} & .65 & .55\\ 
        Property    & .18\small{$\pm$.02} & .12 & .36 & .42\small{$\pm$.03} & .29 & .79 & .33\small{$\pm$.09} & .46 & .26 & \textbf{.52}\small{$\pm$.04} & .53 & .51 \\ 
        Pivot    & .07\small{$\pm$.01} & .04 & .31 & .22\small{$\pm$.03} & .13 & .67 & .18\small{$\pm$.04} & .29 & .13 & \textbf{.32}\small{$\pm$.04} & .45 & .25\\\hline 
        \end{tabular}\hspace*{-1cm}\vspace{-0.1cm}
    \caption{F1, precision and recall for the target class in the automatic detection of argument components in tweets. Each experiment was carried out with three randomized partitions, the mean and standard deviation of the F1 are presented. Best results for F1 for each category are highlighted in boldface.} 
    \vspace{-0.1cm}
    \label{tab:evaluation_results}
\end{table*}

Table \ref{tab:evaluation_results} shows the F1 scores of predictions done by our models on the task of identifying argumentative components and information on the tweets. For the BiLSTM+CRF architecture, some experiments are currently underway and cannot be reported at this moment. It can be seen that the logistic regression baselines consistently perform worse than neural architectures. However, when contextual embeddings are used, the performance of the logistic regression approaches that of neural architectures, even if it is lower. Thus, it seems that the task is systematically complex enough that it requires more expressivity than a simple linear classifier. RoBERTa systematically achieves better performance than the BiLSTM+CRF architecture, except for Conclusion and Argumentative vs. Non-Argumentative, where the BiLSTM-CRF model performs better. 

Across categories, we can see that identifying Justifications is the task with the best performance. However, identifying Conclusions seems much more trying, with 20 points less for F1. The rest of aspects of argumentation also present performance around .50 F1, with Collective as somewhat easier to identify. Pivots and Types of Justification are the categories with worse performance.

Results also show that providing the models with the information of relevant previously annotated elements improves performance for Collective and Types of Justification and Conclusion in more than 20 points, but it does not improve performance for Pivot. For Collective, we noticed that models without this information correctly mark targets of hate messages on tweets they shouldn't because they don't have an explicit Property and this doesn't happen when adding information about the Property.



\begin{table}[h]
\centering
\begin{tabular}{l|c|c} 

& LR w/embed & RoBERTa \\
        \hline
        
        Type Justif & .48\small{$\pm$.06} & \textbf{.64}\small{$\pm$.06} \\
        Type Conc & .60\small{$\pm$.03} & \textbf{.73}\small{$\pm$.03} \\
        Collective & .50\small{$\pm$.01} & \textbf{.78}\small{$\pm$.03} \\
        Pivot  & .22\small{$\pm$.02} & \textbf{.25}\small{$\pm$.01}  \\    
        \hline
    \end{tabular}
    \caption{Results for identification of components with providing relevant previously annotated information.}
    \label{tab:results_with_preannotation}
\end{table}

These results show that identifying some of the argumentative information can be addressed using standard state-of-the-art models. For most of the components, the F1 score is equal or even higher than the one calculated for the agreement between the two annotators: considering that in many cases there is more than one possible correct answer, this means that a language model can potentially achieve a performance similar to a human.

\section{Discussion}

In this work, we present the Argumentation Structure For Counter-Narrative Generation (ASFoCoNG) corpus and annotation guidelines. We have adapted the analytic approach proposed by~\citet{Wagemans2016ConstructingAP} 
to enrich a reference dataset for hate speech with argumentative information, aimed to improve the task of counter-narrative generation. The published dataset is a contribution to the existing corpora of counter-narratives generation for hate-speech. 




We addressed our research question about what aspects of argumentative information can be recognized systematically by humans, what aspects can be automated and to which extent. We have found an inter-annotator agreement above $\kappa = .5$ for almost all categories, and above $\kappa = .7$ for recognizing Argumentative tweets, Justifications and Conclusions. Considering we are dealing with user-generated text, we find this a very hopeful scenario. 
We have found that machines perform worse than humans in recognizing argumentative tweets, the Pivot and Conclusions, but comparably for the rest of the aspects.


With respect to Pivots, which are a keystone of our annotation, we will be exploring similarity-based metrics, to assess the semantic equivalence of human and automatic differences. We will explore metrics based on word similarity that can assign a score to a prediction rather than a binary classification in order to better represent cases where two examples have different words with similar meaning labeled and are currently considered as mismatching.

We are also working on training models for jointly predicting complementary aspects of argumentative information at the same time, namely, the Collective and Property, the Justification and Conclusion and the premises along with its type. We are also working on multilingual experiments using the English and Spanish portions of the dataset.

Finally, we are working on training models for counter-narratives generation using our dataset, with and without the argumentative information labeled. We want to assess to what extent does this information help on the task. We will compare results using different proportions and combinations of types of information to see if some of them help more than others and what types of counter-narrative are affected most. We also plan to enhance our dataset semiautomatically.

\section{Bibliographical References}\label{reference}

\bibliography{custom,HSCN}
\bibliographystyle{acl_natbib}

\appendix

\section{Examples of Disagreement between annotators}
\label{app:examples}
\begin{figure*}
    \centering
    
    \includegraphics[width=\linewidth]{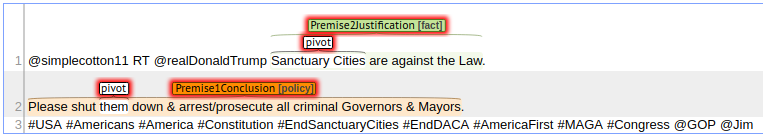}\\
    \includegraphics[width=\linewidth]{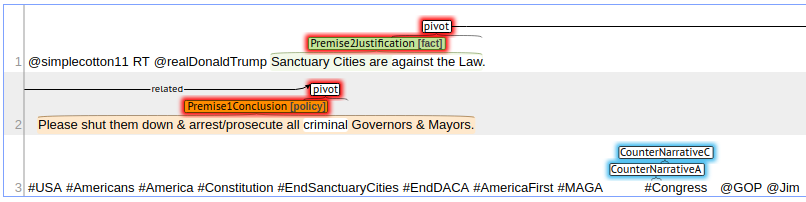}
    \caption{Disagreement between annotators concerning the pivot between a justification and a conclusion. The annotator above believes the pivot is between the words \textit{Sanctuary Cities -- them}, while the annotator below considers the pivot \textit{against the law -- criminal}.}
    \label{fig:disagreement_pivot}
\end{figure*}

Figure~\ref{fig:disagreement_pivot} shows an example of disagreement between annotators concerning the pivot between a justification and a conclusion. One of the annotators considered that the transfer of reason between the two components is centered on \textit{Sanctuary Cities}, while the other considered that criminality is the pivot between justification and conclusion. Both are arguably valid interpretations of the tweet, and both can be useful to construct a counter-narrative, for example, to invalidate the relation between premises. In this case, disagreements seem to arise from differences in interpretation that are inherent to the vagueness of the problem itself.
    
\begin{figure*}
    \centering
    \includegraphics[width=\linewidth]{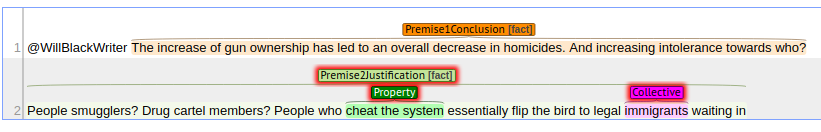}\\
    \includegraphics[width=\linewidth]{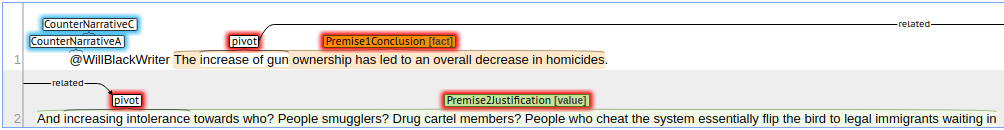}
    \caption{Disagreement between annotators concerning the type of justification. The annotator above believes the justification is a fact, while the annotator below considers the speaker is stating an opinion. Note that there is also disagreement concerning the pivot, the collective and the property.}
    \label{fig:disagreement_type_justification}
\end{figure*}

Annotators also disagree to assign a category of \textit{policy}, \textit{fact} or \textit{value} to justifications, but not to conclusions. An example of disagreement can be seen in Figure~\ref{fig:disagreement_type_justification}. In this case, a long justification includes heterogeneous text, that can be considered both opinion or statement of fact. We will be carrying out more pairwise annotations to gain further insight on the reasons for such low agreement, and to check that it is not due solely to chance, because the current dataset to inspect inter-annotator agreement is small.

\section{Hyperparameters used for model training}

Tables \ref{tab:hyperparameters1} and \ref{tab:hyperparameters2} shows the hyperparameters used for training the best performing models. For RoBERTa, we report the learning rate, for Logistic Regressions, the inverse of the regularization penalty and for BiLSTM we report the last layer used (CRF or Softmax) and the word embbedings model (Komninos, Fastext or Elmo). All RoBERTa models were trained using a batch size of 32 and a weight decay of 0.01. All Logistic Regressions were trained using an L2 penalization, a liblinear solver and a 'balanced' class weight.

\begin{table}[ht!]
\centering
\begin{tabular}{l|c|c|c|c} 

        \hline
                       & LR & LR & BiLSTM & Roberta\\
                       &    &+Embed& +CRF & Base \\
        \hline
        Arg vs Non-arg & 1.0 & 1.0 & CRF/Komninos & 5e-05 \\
        Justification & 1.0 & 0.1 & Softmax/Elmo & 8e-05 \\
        Conclusion & 0.1 & 0.1 & Softmax/Elmo & 5e-05 \\
        Type Justif & 1.0 & 0.1 & - & 5e-05 \\
        Type Conc & 1.0 & 0.5 & - & 8e-05 \\
        Collective & 1.0 & 0.5 & CRF/Komninos & 2e-05 \\
        Property & 0.5 & 0.1 & Softmax/Elmo & 1e-04 \\
        Pivot  & 1.0 & 1.0 & Softmax/Elmo & 2e-05 \\

    \end{tabular}
    
    \caption{Hyperparameters used on the best performing models for each experiment without extra information}
    \label{tab:hyperparameters1}
\end{table}

\begin{table}[ht!]
\centering
    \begin{tabular}{l|c|c}
    
                           & LR & Roberta\\
                       &+Embed & Base \\
            \hline
        
        Type Justif  & 0.1 & 1e-04\\
        Type Conc  & 1.0 & 5e-05\\
        Collective  & 1.0  & 5e-05\\
        Pivot  & 0.2 & 2e-05\\
        
        \hline
    \end{tabular}
        \caption{Hyperparameters used on the best performing models for each experiment adding pre-annotated information}
    \label{tab:hyperparameters2}
\end{table}


\end{document}